\documentclass[pdflatex,sn-nature]{sn-jnl}

\usepackage{amsmath,amssymb,amsfonts}
\usepackage{graphicx}
\usepackage{booktabs}
\usepackage{multirow}
\usepackage{tabularx}
\usepackage{array}
\usepackage{placeins}

\let\oldFloatBarrier\FloatBarrier
\renewcommand{\FloatBarrier}{%
    \oldFloatBarrier
    \vspace{-0.5\baselineskip}%
}
\usepackage[caption=false]{subfig}
\usepackage{tikz}
\usetikzlibrary{
    arrows.meta,
    positioning,
    calc,
    fit,
    backgrounds,
    shapes.geometric
}

\begin{document}

\title{TWIG: A Time-Causal Wavelet Operator for Autoregressive Forecasting on Irregular Graphs}

\author[1, 3]{\fnm{Subashree} \sur{Venkatasubramanian}}

\author[2]{\fnm{David A.} \sur{Barajas-Solano}}

\author[1]{\fnm{Chuyang} \sur{Liu}}

\author[3]{\fnm{Daniel M.} \sur{Tartakovsky}}

\author*[1]{\fnm{Dipankar} \sur{Dwivedi}}
\email{ddwivedi@lbl.gov} 

\affil[1]{
    \orgname{Lawrence Berkeley National Laboratory},
    \orgaddress{
        \city{Berkeley},
        \state{CA},
        \country{USA}
    }
}

\affil[2]{
    \orgname{Pacific Northwest National Laboratory},
    \orgaddress{
        \city{Richland},
        \state{WA},
        \country{USA}
    }
}

\affil[3]{
    \orgname{Stanford University},
    \orgaddress{
        \city{Stanford},
        \state{CA},
        \country{USA}
    }
}

\abstract{
We introduce TWIG (Time-Causal Wavelet Operator for Irregular Graphs), a graph-native neural operator for autoregressive surrogate modeling on static irregular graphs. TWIG transforms each node history into causal multiscale temporal features that separate recent variation from progressively slower memory components, then propagates these features through graph-wavelet operator blocks with gated pointwise channel mixing. The architecture is causal by construction and designed for closed-loop forecasting, where predictions are recursively reused as future inputs. We evaluate TWIG on three irregular-domain forecasting problems spanning regional diffusion, three-dimensional subsurface hydrology, and aerodynamic flow, with graphs ranging from 400 to 5,233 nodes and model capacities from approximately 70k to 10M parameters. TWIG achieves the lowest aggregate rollout errors on the subsurface-hydrology and regional-diffusion benchmarks and ranks second on the 10M-parameter aerodynamic-flow benchmark, behind the GPS Transformer. Across all three settings, TWIG consistently outperforms the corresponding non-time-causal Graph WNO baseline. These results demonstrate that TWIG provides an effective and scalable
approach to stable autoregressive forecasting of dynamical fields on
irregular graphs.
}

\maketitle

High-fidelity simulation is a primary tool for scientific discovery, enabling researchers to investigate physical processes that cannot be easily measured in the lab or field.
Through numerical simulation, we can model complex dynamics such as coastal state responses, subsurface contaminant transport, or regional spread of an infectious disease. Yet the same simulations that make these questions accessible often carry high computational costs. Realistic physical models must resolve geometry, heterogeneity, boundary conditions, and coupled processes over many time steps. As a result, a single forward solve may be manageable, but the repeated solves required for long-term behavioral observation in tasks such as uncertainty quantification or data assimilation can quickly become a bottleneck \cite{karniadakis2021physics,kovachki2023neural}. This challenge is well illustrated in subsurface modeling, where recent surrogate approaches have reduced PFLOTRAN-based seawater-intrusion simulations from hours to seconds per prediction, demonstrating the substantial computational gains available from learned approximations \cite{jiang2024geofusehighefficiencysurrogatemodel}.

Designing such surrogates becomes more difficult when the underlying physical domain is not naturally represented by a regular grid. Subsurface systems may contain complex geological structures, coastlines, and engineered infrastructure, while epidemiological dynamics may evolve over regional interaction networks defined by transportation or administrative connectivity. Such simulations often rely on unstructured meshes that conform
to complex geometries, with states represented on irregularly distributed
nodes connected by mesh or physical adjacency
\cite{shuman2013emerging,pfaff2021learning,wu2022learning,
cao2023efficient,meshbook}. Useful surrogates for these systems must preserve
this underlying geometry, communicate over local and long-range graph
structure, and remain stable when predictions are recursively reused as inputs.

Learned surrogates approximate reusable solution operators across parameter settings, initial conditions, and forecast states  \cite{karniadakis2021physics,lu2021learning,li2021fourier,kovachki2023neural}. They have shown promise in computational fluid dynamics and subsurface data-assimilation workflows
\cite{kochkov2021machine,tang2020deep,li2023gino, venkatasubramanian2024variationalencoderdecoderslearninglatent}, while learned weather forecasting systems such as GraphCast and ECMWF's AIFS demonstrate that data-driven models can produce large-scale forecasts at a fraction of the computational cost of traditional numerical pipelines \cite{lam2023graphcast,lang2024aifs,moldovan2026aifs}. These advances motivate models that learn dynamical evolution directly on a physical system's native discretization for rapid scenario analysis and ensemble forecasting.

Most foundational neural operators, however, were developed for structured Euclidean domains. Fourier neural operators (FNOs) exploit global spectral convolution on regular grids \cite{li2021fourier}, while wavelet neural operators (WNOs) provide multiresolution spatial representations but are likewise most naturally formulated on tensorized domains
\cite{tripura2023wavelet}. Graph neural networks (GNNs) and mesh-based simulators instead operate naturally on irregular discretizations \cite{gilmer2017neural,sanchezgonzalez2020learning,pfaff2021learning, brandstetter2022message}, although local message passing can require many layers to communicate over long spatial distances and may compress distant information through graph bottlenecks
\cite{li2020multipole,alon2021bottleneck,topping2022understanding, dwivedi2022long}. Attention-based GNNs and graph transformers provide non-local alternatives with different scaling and capacity trade-offs \cite{velickovic2018graph,brody2022how,kreuzer2021rethinking,rampasek2022recipe,alkin2024upt}, while geometry-informed and graph-based neural operators extend operator learning to meshes, point clouds and other irregular representations \cite{li2020multipole,li2023gino,li2023fouriergeometry,li2023gino}. Graph-native surrogates have likewise been applied to domains including urban water systems and disease spread \cite{garzon2022machine,panagopoulos2021transfer}. These developments address many of the spatial challenges of learning on irregular domains, but they do not by themselves resolve the temporal instability that arises when learned evolution operators are deployed over long autoregressive rollouts \cite{mccabe2023stability,koehler2024apebench}.

During training, a surrogate often sees clean histories drawn from the data distribution; during deployment, it must instead continue from histories containing its own previous errors. Small mistakes can therefore compound, shift the input distribution and eventually dominate the forecast \cite{bengio2015scheduled,vlachas2024learning}. This problem is particularly important in multiscale physical systems, where fast transients, intermediate transport and slow drift may coexist within the same node history. Simply concatenating several past states as input channels provides temporal context, but does not impose any structure on that context. The model must therefore infer from raw history alone which variations are transient, which persist across scales and which are most relevant to future evolution.

\begin{figure*}[!htbp]
    \centering

    \subfloat[
        TWIG combines time-causal multiscale temporal features with structural
        positional encodings, projects the resulting representation into a
        hidden space, processes it through \(L\) GraphWNO blocks, and decodes
        a future-state window. The predicted window is recursively reused as
        input during block-autoregressive rollout.
        \label{fig:twig-architecture}
    ]{
        \includegraphics[
            width=0.85\textwidth
        ]{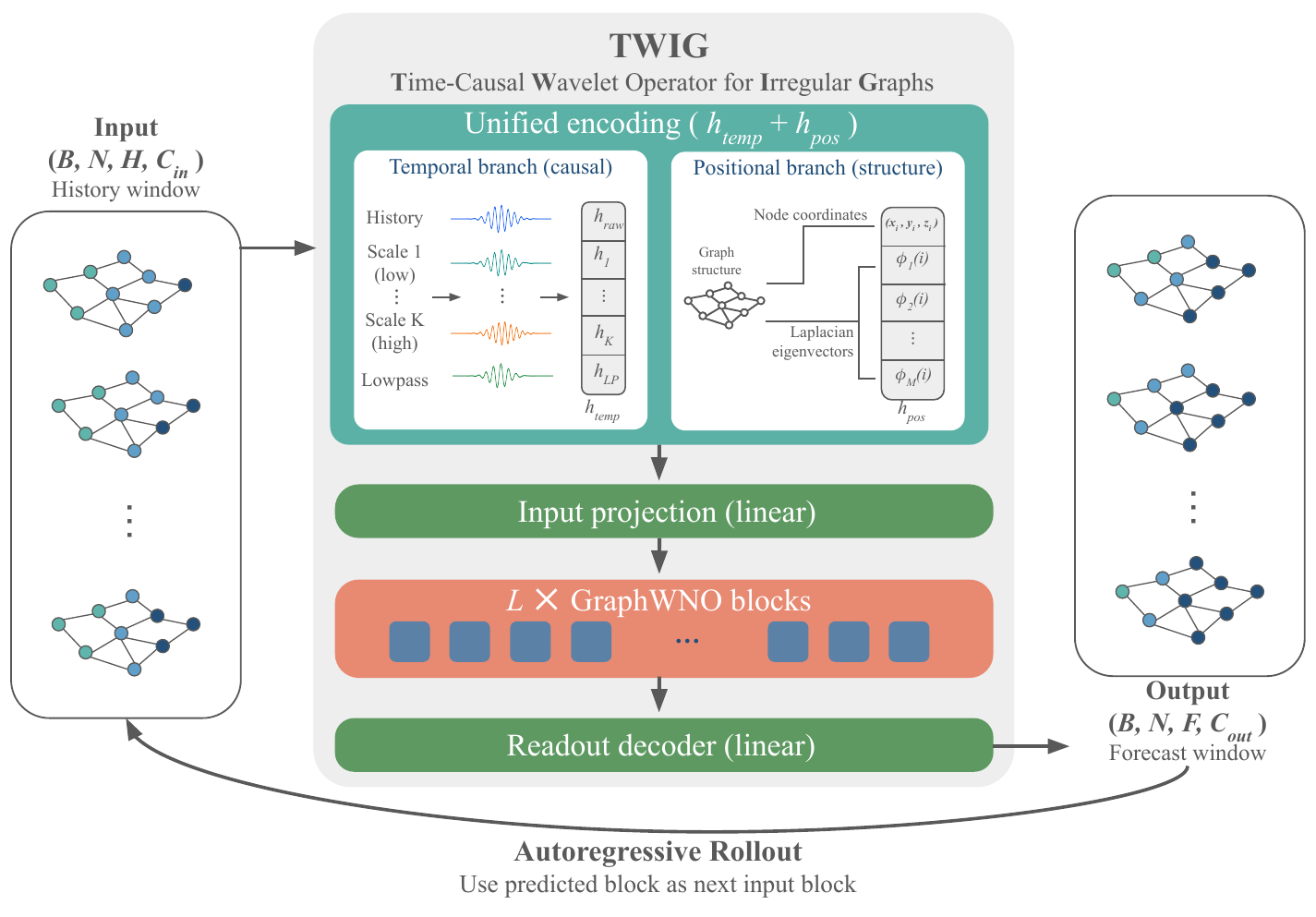}
    }

    \par\vspace{0.3em}

    \subfloat[
        Each Graph WNO block applies multiscale graph-wavelet spatial
        processing followed by pointwise SwiGLU channel mixing, with residual
        connections around both sublayers.
        \label{fig:graphwno-block}
    ]{
        \includegraphics[
            width=0.85\textwidth
        ]{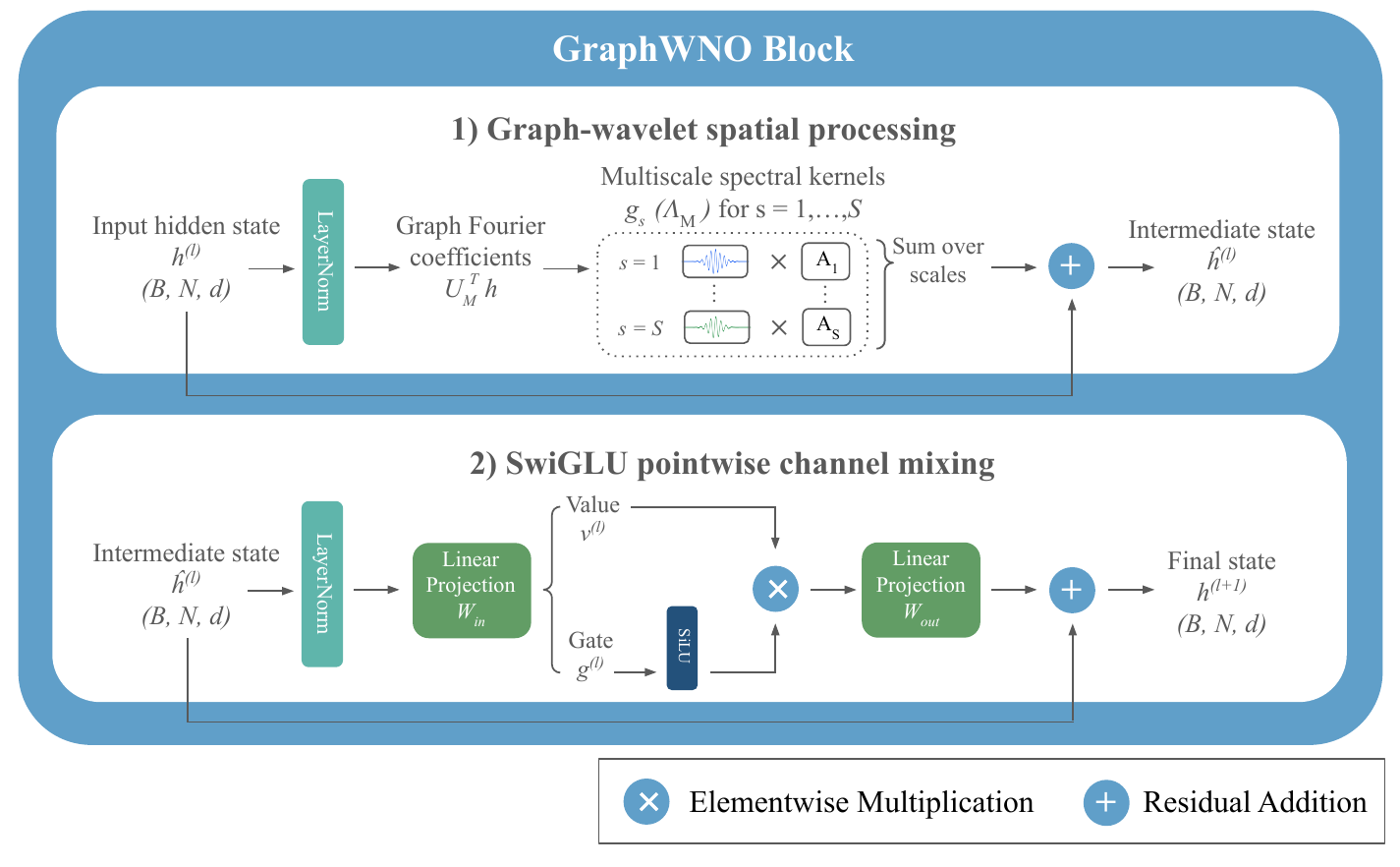}
    }

    \caption{
        \textbf{Architecture of TWIG (Time-Causal Wavelet Operator for Irregular Graphs).}
        \textbf{a}, Overall TWIG architecture and block-autoregressive forecasting
        procedure.
        \textbf{b}, Structure of the GraphWNO block used for spatial propagation
        and pointwise channel mixing.
    }
    \label{fig:twig-overview}
\end{figure*}

We introduce TWIG (\textbf{T}ime-Causal \textbf{W}avelet Operator for \textbf{I}rregular \textbf{G}raphs), a graph-native neural operator designed around this separation of temporal and spatial roles. The architecture is summarized in Figure~\ref{fig:twig-architecture}. Its temporal component converts each node's finite history into a causal multiscale representation consisting of the current state, temporal residuals at progressively slower memory scales, and a low-pass memory component. This construction is inspired by time-causal multiscale representations, in which features at time \(t\) depend only on observations available at or before \(t\) \cite{lindeberg2025timecausal}. The spatial component then propagates these features through graph-wavelet operator blocks, using a graph-Laplacian basis to communicate information across the irregular domain \cite{hammond2011wavelets,shuman2013emerging,xu2019graphwavelet}. A static positional-encoding branch supplies structural information through node coordinates and graph-spectral features \cite{kreuzer2021rethinking,rampasek2022recipe}, while a pointwise SwiGLU channel mixer provides gated feature interactions within each node \cite{shazeer2020glu}. TWIG is therefore causal in time, graph-native in space, and designed for closed-loop autoregressive forecasting rather than one-step regression.

\section{Results}

We evaluate TWIG on three irregular-domain forecasting benchmarks spanning
subsurface hydrology, regional diffusion, and aerodynamic flow. The
experiments cover model capacities from approximately \(70\)k to \(10\)M
parameters and autoregressive forecast horizons from 60 to 180 steps. Within
each benchmark, all models use the same scenario-level data splits, training
configuration, and block-autoregressive evaluation protocol, and reported
results are averaged over three independently trained runs. Dataset
construction, optimization settings, and complete model configurations are
provided in Methods and the Appendix.

\subsection{TWIG framework}
\label{sec:twig-framework}

TWIG separates temporal representation from spatial propagation
(Fig.~\ref{fig:twig-overview}). At each graph node, the available history
is mapped to a causal multiscale representation containing the current state,
\(K\) temporal residual bands, and a slowly varying memory component. Static
structural information from node coordinates and graph-Laplacian eigenvectors
is incorporated through a parallel positional branch. The resulting features
are then projected into a hidden representation and processed by \(L\)
GraphWNO blocks.

Each GraphWNO block combines multiscale graph-wavelet propagation with
pointwise SwiGLU channel mixing (Fig.~\ref{fig:graphwno-block}). The decoder
jointly predicts a block of future graph states, which is recursively returned
as input during long-horizon rollout. Consequently, forecast states beyond
the initial context are generated entirely from previous model predictions.
The full mathematical formulation is given in Methods.

\subsection{Long-horizon forecasting across irregular domains}
\label{sec:benchmark-results}

\begin{figure*}[!htbp]
    \centering
    \includegraphics[
        width=\textwidth
    ]{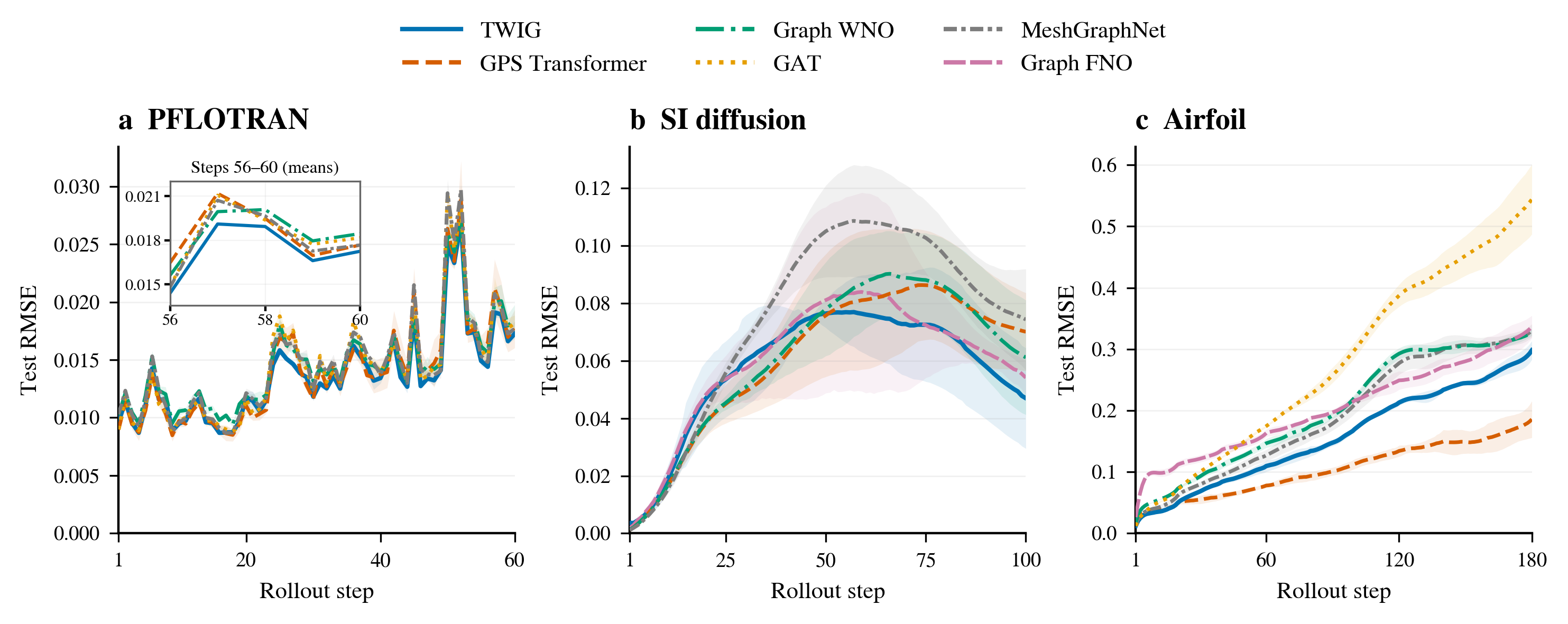}
    \caption{
        \textbf{Autoregressive forecasting across three irregular-domain
        benchmarks.}
        \textbf{a}, PFLOTRAN test RMSE over a 60-step rollout; the inset
        enlarges the final five forecast steps to distinguish the leading
        models.
        \textbf{b}, SI-diffusion test RMSE over a 100-step rollout.
        \textbf{c}, Airfoil test RMSE over a 180-step rollout.
        Curves denote the mean across three independently trained runs and
        shaded regions denote one standard deviation. Selected competitive
        baselines and the non-time-causal Graph WNO control are shown for
        visual clarity; complete benchmark comparisons are reported in the
        Appendix.
    }
    \label{fig:rollout-summary}
\end{figure*}

We compare TWIG with capacity-matched recurrent, message-passing,
graph-attention, Transformer, and spectral neural-operator baselines.
Figure~\ref{fig:rollout-summary} summarizes autoregressive error growth across
the three benchmarks for TWIG and a representative set of the strongest
competing models. Complete numerical results and comparisons with all
baselines are provided in Appendix
Tables~\ref{tab:pflotran-rollout},
\ref{tab:si-comparison}, and
\ref{tab:Airfoil-comparison}.

The three benchmarks exhibit distinct patterns of autoregressive error
accumulation. PFLOTRAN remains comparatively stable over the 60-step rollout,
with intermittent increases in error shared across the leading models
(Fig.~\ref{fig:rollout-summary}a). SI diffusion exhibits a broader rise and
subsequent decline in error over the 100-step forecast
(Fig.~\ref{fig:rollout-summary}b), whereas errors on Airfoil grow more
persistently over the substantially longer 180-step rollout
(Fig.~\ref{fig:rollout-summary}c). Against these different error-growth
patterns, TWIG achieves the lowest aggregate rollout error on PFLOTRAN and
SI diffusion and ranks second behind the GPS Transformer on Airfoil.

On PFLOTRAN, the leading models remain closely grouped across much of the
rollout, although their errors separate toward the end of the forecast
(Fig.~\ref{fig:rollout-summary}a). TWIG achieves the lowest mean rollout
and final-timestep RMSE, with the GPS Transformer providing the strongest
competing result. Relative to the non-time-causal Graph WNO control, TWIG
reduces mean rollout RMSE by \(6.1\%\) and final-timestep RMSE by \(6.5\%\).
The enlarged final five steps further show that this advantage is retained
through the end of the autoregressive horizon.

A different error profile emerges on SI diffusion
(Fig.~\ref{fig:rollout-summary}b). Errors increase through approximately the
middle of the rollout before decreasing toward the final state, with greater
between-run variability than on PFLOTRAN. TWIG nevertheless achieves the
lowest aggregate rollout and final-timestep errors. Graph FNO provides the
strongest competing aggregate result, while TWIG reduces mean rollout RMSE
by \(6.6\%\) and final-timestep RMSE by \(23.2\%\) relative to Graph WNO.
Thus, the improvement over the non-time-causal graph-wavelet control persists
despite substantially different temporal error dynamics.

Airfoil provides the clearest contrast among the three benchmarks
(Fig.~\ref{fig:rollout-summary}c). Error grows steadily over the 180-step
forecast for all models, but the GPS Transformer accumulates error at a
substantially lower rate and achieves the strongest overall performance.
TWIG ranks second, while remaining ahead of MeshGraphNet, Graph FNO, and
Graph WNO throughout most of the rollout. Relative to the best tested
Graph WNO configuration, TWIG reduces mean rollout RMSE by \(22.3\%\).
The result indicates that the advantage of time-causal processing over the
non-time-causal graph-wavelet control persists at approximately \(10\)M
parameters, although global attention is particularly effective in this
larger aerodynamic-flow setting.

\paragraph*{Spatial structure of long-horizon forecast error}
\label{sec:spatial-error-analysis}

\begin{figure*}[!htbp]
\centering

\subfloat[
    PFLOTRAN final-timestep pressure error.
    \label{fig:spatial-errors-pflotran}
]{
    \includegraphics[
        width=0.98\textwidth
    ]{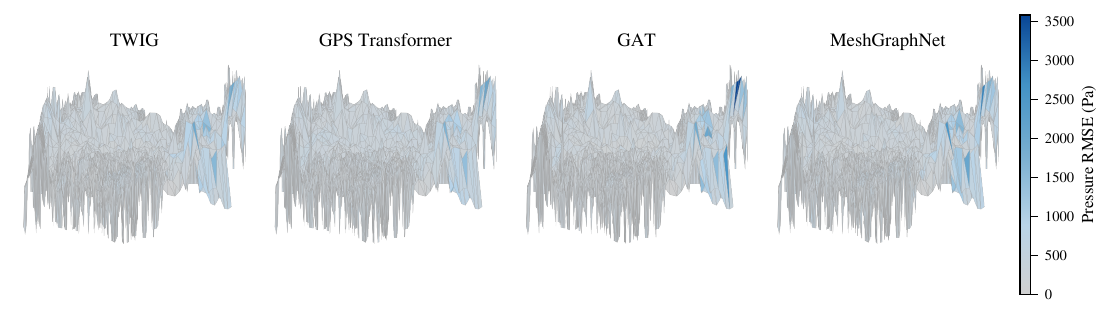}
}

\par\vspace{0.5em}

\subfloat[
    SI-diffusion final-timestep infected-fraction error.
    \label{fig:spatial-errors-si}
]{
    \includegraphics[
        width=0.98\textwidth
    ]{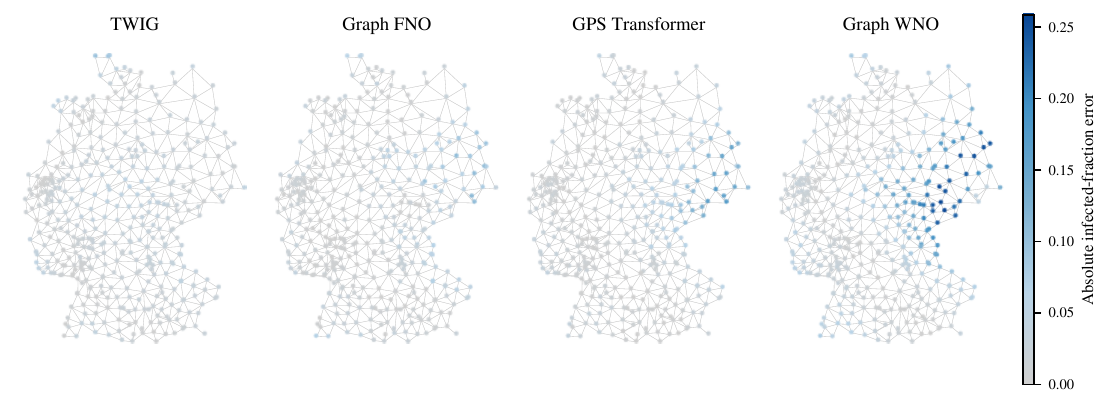}
}

\par\vspace{0.5em}

\subfloat[
    Airfoil final-timestep velocity error.
    \label{fig:spatial-errors-airfoil}
]{
    \includegraphics[
        width=0.98\textwidth
    ]{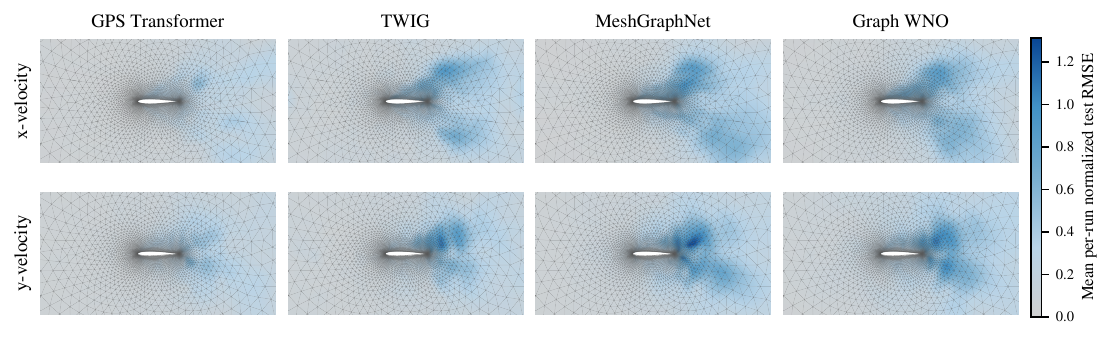}
}

\caption{
\textbf{Spatial distribution of final-timestep forecast error across the
three benchmarks.}
\textbf{a}, Nodewise PFLOTRAN pressure RMSE at forecast step 60.
\textbf{b}, Nodewise absolute infected-fraction error for SI diffusion at
forecast step 100.
\textbf{c}, Nodewise normalized error for the Airfoil \(x\)- and
\(y\)-velocity components at forecast step 180.
Each field is computed over the complete held-out test set and averaged over
the three independently trained model seeds. All models within a panel use a
common color scale, allowing the spatial extent and concentration of forecast
error to be compared directly.
}
\label{fig:spatial-errors}
\end{figure*}

Aggregate rollout metrics do not reveal where forecast errors accumulate over
the physical domain. We therefore examine the spatial distribution of error
at the final forecast state for the leading models
(Fig.~\ref{fig:spatial-errors}). Unlike a single illustrative trajectory,
each visualization is computed over the complete held-out test set and
averaged across the three independently trained model seeds.

On PFLOTRAN, final-timestep pressure error is concentrated within a limited
portion of the subsurface graph rather than being distributed uniformly
throughout the domain (Fig.~\ref{fig:spatial-errors-pflotran}). TWIG
maintains comparatively small errors across most of the graph, while the
competing models exhibit larger localized error concentrations in regions
where forecast uncertainty accumulates during rollout.

The SI-diffusion error fields show a similarly structured spatial pattern
(Fig.~\ref{fig:spatial-errors-si}). TWIG maintains relatively low
infected-fraction error across most of the regional graph, whereas Graph WNO
develops substantially larger errors over a concentrated group of regions.
Graph FNO and the GPS Transformer lie between these behaviors, consistent with
their aggregate rollout performance.

For Airfoil, forecast error is concentrated around and downstream of the
airfoil, where the velocity field contains the strongest spatial structure
(Fig.~\ref{fig:spatial-errors-airfoil}). The GPS Transformer produces the
smallest velocity errors in this region, consistent with its lower aggregate
rollout RMSE. TWIG remains substantially closer to the GPS Transformer
than MeshGraphNet or Graph WNO, while the latter models exhibit broader and
stronger error concentrations in the near-wake region.

\subsection{Multiscale representation and scale sensitivity}
\label{sec:scale-ablations}

\begin{figure*}[!htbp]
    \centering
    \includegraphics[
        width=0.75\textwidth
    ]{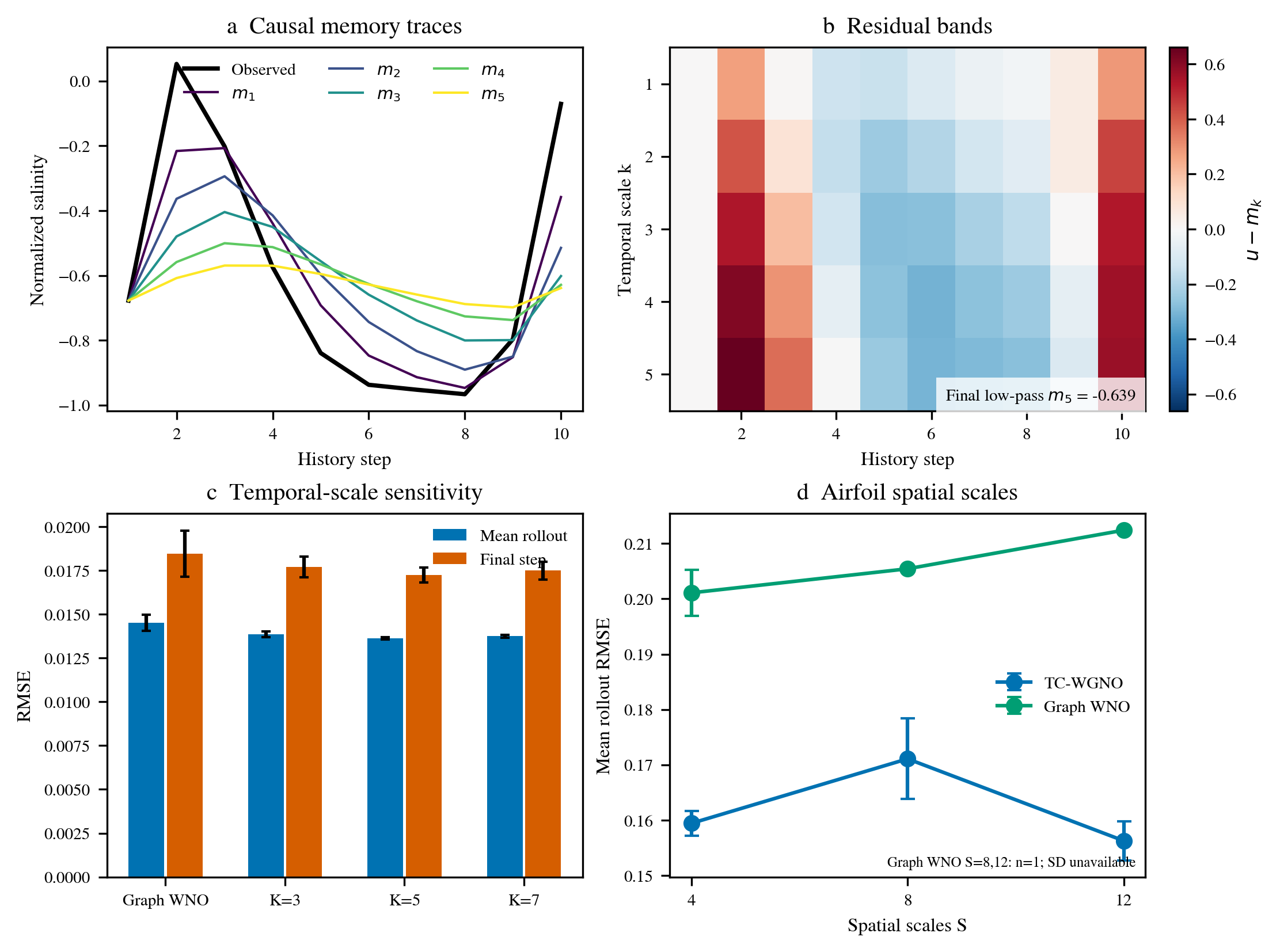}
    \caption{
    \textbf{Time-causal multiscale representation and sensitivity to
    temporal and spatial resolution.}
    \textbf{a}, Causal memory traces for a representative PFLOTRAN salinity
    history using \(K=5\), showing progressively slower responses across
    temporal scales.
    \textbf{b}, Corresponding residual features
    \(u^{(\tau)}-m_k^{(\tau)}\), which encode deviations from memories with
    different temporal spans.
    \textbf{c}, PFLOTRAN mean rollout and final-timestep RMSE for Graph WNO
    and TWIG with \(K=3,5,7\).
    \textbf{d}, Airfoil mean rollout RMSE as a function of the number \(S\)
    of graph-wavelet spatial scales for TWIG and Graph WNO.
    Error bars denote one standard deviation where multiple independently
    trained runs are available.
    }
    \label{fig:multiscale-analysis}
\end{figure*}

Figure~\ref{fig:multiscale-analysis} connects the temporal representation
constructed by TWIG with its sensitivity to temporal and spatial
multiscale resolution. For a representative PFLOTRAN salinity history, the
causal memory traces become progressively smoother with increasing temporal
scale (Fig.~\ref{fig:multiscale-analysis}a). Short-memory states respond
rapidly to recent changes, whereas longer memories preserve information over
a larger fraction of the observed history. The corresponding residual
features \(u^{(\tau)}-m_k^{(\tau)}\) therefore encode the same trajectory
relative to distinct temporal memory scales
(Fig.~\ref{fig:multiscale-analysis}b).

The forecasting improvement is not restricted to a single temporal
resolution. TWIG with \(K=3\), \(5\), and \(7\) outperforms plain Graph
WNO on PFLOTRAN in both mean rollout and final-timestep RMSE
(Fig.~\ref{fig:multiscale-analysis}c). The \(K=5\) configuration performs
best, while the differences among the three time-causal variants remain
comparatively modest. This indicates that the improvement is associated with
the introduction of causal multiscale temporal structure rather than a
narrowly selected value of \(K\).

Spatial-scale sensitivity on Airfoil is likewise non-monotonic
(Fig.~\ref{fig:multiscale-analysis}d). For TWIG, increasing the number of
graph-wavelet scales from \(S=4\) to \(S=8\) increases rollout error, while
\(S=12\) produces the lowest error of the three tested configurations.
Increasing spatial multiscale resolution therefore does not uniformly improve
forecasting accuracy. Across the available matched-scale comparisons,
TWIG remains below Graph WNO, indicating that its advantage is not
restricted to a single graph-wavelet scale choice.

\subsection{Robustness to perturbed initial histories}
\label{sec:pflotran-noise}

Finally, we test whether the long-horizon advantage persists when the
observed history used to initialize the forecast is corrupted at inference
time. The clean-trained PFLOTRAN models are evaluated after independently
perturbing the ten states in the initial history by a fixed fraction of each
channel's training-set standard deviation, while the subsequent reference
trajectory remains unchanged.

\begin{figure*}[!htbp]
    \centering
    \includegraphics[
        width=\textwidth
    ]{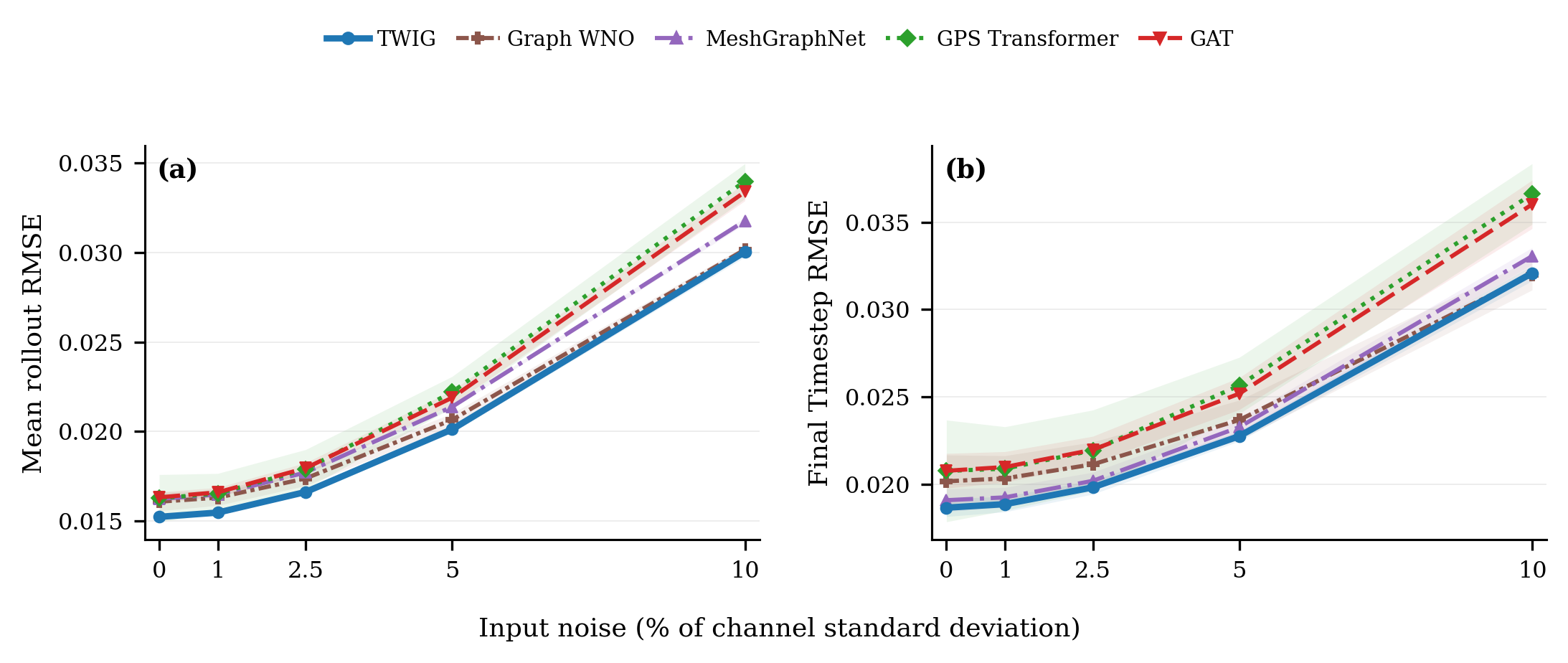}
    \caption{
        \textbf{Robustness of PFLOTRAN forecasts to perturbations of the
        initial history.}
        The initial ten-state context is perturbed by the indicated
        percentage of each channel's training-set standard deviation.
        The left panel reports mean RMSE over the \(60\)-step rollout and
        the right panel reports final-timestep RMSE. Models are trained only
        on clean data. Curves report means across three independently trained
        runs.
    }
    \label{fig:pflotran-noise-robustness}
\end{figure*}

Forecast error increases with perturbation magnitude for every model
(Fig.~\ref{fig:pflotran-noise-robustness}), but TWIG retains the lowest
mean rollout and final-timestep RMSE across all tested noise levels. The
advantage therefore persists when the autoregressive rollout begins from a
moderately shifted input context rather than only from clean simulation
states.

\section{Discussion}

Across the three benchmarks, the results indicate that explicitly organizing temporal history before spatial graph processing can improve autoregressive forecasting on irregular domains, as TWIG achieves the lowest aggregate rollout errors on PFLOTRAN and SI diffusion and consistently improves over the non-time-causal Graph WNO comparison across all three datasets. This is consistent with prior work identifying recursive error accumulation and distribution shift as central challenges in learned dynamical forecasting
\cite{mccabe2023stability,koehler2024apebench,vlachas2024learning}. TWIG approaches this problem architecturally by separating recent variation from progressively slower memory components before graph-wavelet propagation, rather than requiring the spatial processor to infer temporal structure directly from concatenated history states. This behavior persists across graph sizes ranging from 400 to more than 5,000 nodes and model capacities from approximately \(70\)k to \(10\)M parameters. The results therefore suggest that the benefit of the time-causal representation is not confined to a particular graph size or application.

The Airfoil benchmark also qualifies this result. At the \(10\)M-parameter scale, the GPS Transformer outperforms TWIG, while TWIG remains ahead of the message-passing, graph-attention, recurrent, Graph FNO, and Graph WNO baselines. The strong Airfoil performance of GPS is consistent with prior evidence that combining local graph processing with global attention can be effective for capturing long-range interactions on graphs
\cite{rampasek2022recipe}. In contrast, TWIG outperforms the GPS Transformer on both PFLOTRAN and SI diffusion, which operate at approximately \(1\)M and \(70\)k parameters and use only 80 and 18 training scenarios,
respectively, compared with 1,000 training trajectories for Airfoil. Although these experiments do not isolate model capacity from dataset size or task difficulty, this pattern suggests that an explicit causal temporal representation may be particularly useful in capacity- and data-constrained
scientific machine-learning settings. TWIG's persistent advantage over Graph WNO further indicates that structured temporal processing remains beneficial when the graph-wavelet architecture is scaled to substantially larger capacity.

The scale and perturbation experiments provide additional insight into this behavior. On PFLOTRAN, each tested time-causal temporal resolution improves over Graph WNO, while the differences among the TWIG configurations are comparatively modest. Similarly, on Airfoil, TWIG outperforms Graph WNO at every matched graph-wavelet resolution. The perturbation experiment further shows that the forecasting advantage is retained when the observed history is corrupted at inference time, rather than appearing only under clean initialization. Together, these findings suggest that the observed improvements are relatively robust to the precise multiscale configuration and to moderate shifts in the initial forecast context.

Several limitations remain. All three benchmarks use fixed graph topologies, so the present experiments do not establish generalization to changing connectivity or unseen meshes, and the capacity-matched baselines do not exhaust the possible performance of each model family. It should also be noted that TWIG replaces the conventional feed-forward mixer with SwiGLU, so comparison with plain Graph WNO does not attribute the full performance difference exclusively to the time-causal encoder. Future work should examine these components independently and extend the approach to dynamic graphs, mesh-transfer settings, and larger scientific surrogate models.
\FloatBarrier

\section{Methods}
\label{sec:methods}

\subsection{Problem formulation and block-autoregressive forecasting}
\label{sec:block-forecasting}

Let \(G=(V,E)\) denote a static graph with \(N=|V|\) nodes. At time \(t\),
the physical state is represented by a matrix
\(U^{(t)}\in\mathbb{R}^{N\times C}\), where \(C\) is the number of predicted
state variables. Given the most recent \(H\) graph states, the forecasting
model predicts the following \(F\) states jointly:
\begin{equation}
\begin{aligned}
\mathcal{H}^{(t)}
    &=
    \left[
        U^{(t-H+1)},\ldots,U^{(t)}
    \right],\\
\widehat{U}^{(t+1:t+F)}
    &=
    f_{\theta}
    \left(
        \mathcal{H}^{(t)},G,a
    \right).
\end{aligned}
\label{eq:block-forecast}
\end{equation}
Here, \(a\) collects static graph information available to the model, such as node coordinates, graph-spectral positional features, edge connectivity, and edge attributes.

Predicting a block rather than a single future state reduces the number of model evaluations required during long rollouts and allows the network to
represent temporal interactions within the forecast window. At rollout block \(b+1\), the model predicts
\begin{equation}
\widehat{\mathcal{B}}^{(b+1)}
=
f_{\theta}
\left(
    \mathcal{H}^{(b)},G,a
\right),
\end{equation}
after which the predicted states are appended to the current history and the
oldest states are discarded so that the context again contains exactly
\(H\) states.

After the initial observed context, no additional ground-truth states are
provided. Rollout performance therefore reflects both direct block-prediction
accuracy and the stability of the learned dynamics under recursively generated
inputs.

The predicted variables differ across the three applications. In the PFLOTRAN
experiments, each node contains normalized liquid pressure and salinity,
\(u_i^{(t)}=[p_i^{(t)},s_i^{(t)}]\). In the SI-diffusion benchmark, the
underlying simulator evolves susceptible and infected populations, while the
forecasting task predicts only the infected-state field \(I^{(t)}\). In the
Airfoil benchmark, each node contains density, pressure, and the two Cartesian
velocity components,
\(u_i^{(t)}=[\rho_i^{(t)},p_i^{(t)},v_{x,i}^{(t)},v_{y,i}^{(t)}]\).

\subsection{Time-Causal Wavelet Operator for Irregular Graphs}
\label{sec:twig}

TWIG consists of three stages. First, a time-causal multiscale encoder
summarizes the input history at several temporal resolutions. Second, a
graph-wavelet operator exchanges information across the spatial graph.
Finally, a pointwise SwiGLU mixer transforms the latent channels independently
at each node. Repeating the graph-wavelet and channel-mixing stages produces
the latent representation used to decode the complete future block.

\subsubsection{Time-causal multiscale temporal encoding}
\label{sec:tc-encoding}

For node \(i\), let
\(\mathcal{H}_{i}^{(t)}
=[u_i^{(t-H+1)},\ldots,u_i^{(t)}]\)
denote the available temporal context. TWIG constructs a collection of
exponentially weighted memory states with different decay rates. For temporal
band \(k\), the recurrence is
\begin{equation}
\begin{aligned}
m_{i,k}^{(t-H+1)}
    &=
    u_i^{(t-H+1)},\\
m_{i,k}^{(\tau)}
    &=
    \alpha_k m_{i,k}^{(\tau-1)}
    +
    (1-\alpha_k)u_i^{(\tau)},
    \qquad
    \tau=t-H+2,\ldots,t.
\end{aligned}
\label{eq:causal-memory}
\end{equation}
Every memory state therefore depends only on observations available at or
before the current time. Small values of \(\alpha_k\) respond quickly to recent
changes, whereas values closer to one retain information over a longer
temporal span.

In the scale-aware construction used in the main TWIG experiments, the
memory spans are determined relative to the available history length rather
than being fixed in absolute time. For \(K\) temporal bands, we set
\begin{equation}
\begin{aligned}
\tau_k
    &=
    H^{\frac{k-1}{K-1}},
    \qquad k=1,\ldots,K,\\
\alpha_k
    &=
    \exp\left(-\frac{1}{\tau_k}\right).
\end{aligned}
\label{eq:scale-aware-temporal-bands}
\end{equation}
The resulting characteristic spans are distributed geometrically between one
time step and the full context length \(H\). Thus, the shortest memories
emphasize rapidly changing behavior, while the slowest memory summarizes
variation across nearly the entire observed window.

The final temporal representation combines the most recent state with
deviations from the multiscale memories and the slowest low-pass component:
\begin{equation}
\Phi_{\mathrm{TC}}
\left(
    \mathcal{H}_{i}^{(t)}
\right)
=
\left[
    u_i^{(t)}
    \,\Vert\,
    u_i^{(t)}-m_{i,1}^{(t)}
    \,\Vert\,
    \cdots
    \,\Vert\,
    u_i^{(t)}-m_{i,K}^{(t)}
    \,\Vert\,
    m_{i,K}^{(t)}
\right].
\label{eq:tc-features}
\end{equation}
The residual channels act as temporal high-pass features at progressively
longer scales, while the final memory preserves the slowly varying component.
This design is inspired by time-causal temporal scale-space
representations~\cite{lindeberg2025timecausal}, but it is not an orthogonal
wavelet transform and does not rely on exact reconstruction or analytic
scale-covariance guarantees.

\subsubsection{Graph-wavelet spatial processing}
\label{sec:graph-wavelet-operator}

Let \(L_G\) be a symmetric graph Laplacian with eigendecomposition
\(L_G=U\Lambda U^{\top}\). We retain the first \(M\) eigenvectors and
eigenvalues, denoted by \(U_M\) and \(\Lambda_M\), and use them to construct a
truncated graph-spectral representation. Before graph processing, the temporal
features are concatenated with fixed node features \(p\) and lifted into a
\(d\)-dimensional hidden space:
\begin{equation}
h^{(0)}
=
P_{\mathrm{in}}
\left(
    \Phi_{\mathrm{TC}}
    \left(
        \mathcal{H}^{(t)}
    \right)
    \,\Vert\,
    p
\right).
\label{eq:twig-input}
\end{equation}

At graph-wavelet scale \(s_r\), a spectral kernel
\(g_{s_r}(\Lambda_M)\) filters the graph Fourier coefficients. The responses
from \(S\) scales are independently transformed in channel space and then
combined:
\begin{equation}
\mathcal{W}_{\theta}(h)
=
\sum_{r=1}^{S}
U_M
\left[
    g_{s_r}(\Lambda_M)
    \odot
    \left(
        U_M^{\top}h
    \right)
\right]
A_r .
\label{eq:graph-wavelet}
\end{equation}
Here, \(A_r\) is a learned channel transformation and
\(\odot\) denotes elementwise multiplication over spectral modes. Spectral
graph wavelets construct scale-dependent filters by applying scaled kernels
to the graph-Laplacian spectrum, providing a multiresolution representation
for signals defined on irregular graphs
\cite{hammond2011wavelets,shuman2013emerging}. Our operator uses this
principle within a learned neural-operator block: responses from several
graph-spectral scales are computed and combined in a single spatial update.
In this sense, Eq.~\eqref{eq:graph-wavelet} adapts the multiscale
wavelet-domain operator principle of WNO~\cite{tripura2023wavelet} from
regular domains to a graph-Laplacian basis.

\subsubsection{SwiGLU pointwise channel mixing}
\label{sec:swiglu-mixer}

Each graph-wavelet update is followed by a pointwise gated channel mixer. The same operation is applied independently at every node, allowing interactions among latent feature channels without modifying the graph connectivity. We use a SwiGLU mixer~\cite{shazeer2020glu} in place of the two-layer Gaussian Error Linear Unit (GELU) feed-forward network used in the plain Graph WNO control.

For block \(\ell\), let \(\operatorname{LN}\) denote layer normalization and \(\operatorname{SiLU}\) denote the Sigmoid Linear Unit activation. The complete residual update is
\begin{equation}
\begin{aligned}
\widetilde{h}^{(\ell)}
    &=
    h^{(\ell)}
    +
    \mathcal{W}_{\theta}^{(\ell)}
    \left(
        \operatorname{LN}
        \left(
            h^{(\ell)}
        \right)
    \right),\\
\left[
    v^{(\ell)},g^{(\ell)}
\right]
    &=
    W_{\mathrm{in}}^{(\ell)}
    \operatorname{LN}
    \left(
        \widetilde{h}^{(\ell)}
    \right)
    +
    b_{\mathrm{in}}^{(\ell)},\\
h^{(\ell+1)}
    &=
    \widetilde{h}^{(\ell)}
    +
    W_{\mathrm{out}}^{(\ell)}
    \left[
        v^{(\ell)}
        \odot
        \operatorname{SiLU}
        \left(
            g^{(\ell)}
        \right)
    \right]
    +
    b_{\mathrm{out}}^{(\ell)}.
\end{aligned}
\label{eq:twig-block}
\end{equation}
The gate \(g^{(\ell)}\) modulates the transformed features before they are
projected back to the latent dimension. To keep the SwiGLU and conventional
feed-forward variants approximately parameter-matched, the SwiGLU hidden
dimension is set to approximately \(4d/3\). A conventional
\(d\rightarrow2d\rightarrow d\) feed-forward network contains approximately
\(4d^2\) weight parameters, whereas a SwiGLU mixer with hidden dimension
\(d_{\mathrm{ff}}\) contains approximately \(3d\,d_{\mathrm{ff}}\).
Setting \(d_{\mathrm{ff}}\approx4d/3\) therefore gives approximately
\(4d^2\) weights in both cases.

After \(L\) graph-wavelet blocks, a linear decoder produces all \(F\) future
states jointly:
\begin{equation}
\widehat{U}^{(t+1:t+F)}
=
P_{\mathrm{out}}
\left(
    h^{(L)}
\right).
\label{eq:twig-output}
\end{equation}

\subsection{Training objective and model selection}
\label{sec:training-objective}

All normalization statistics are estimated using only the training scenarios.
For channel \(c\), the normalized state is
\[
\widetilde{U}_{i,c}^{(t)}
=
\frac{
    U_{i,c}^{(t)}-\mu_c^{\mathrm{train}}
}{
    \sigma_c^{\mathrm{train}}
}.
\]
Channelwise normalization places the predicted variables on comparable
numerical scales so that the optimization objective is not dominated by the
variable with the largest physical magnitude. This is particularly important
for PFLOTRAN, where pressure and salinity have different units and numerical
ranges.

We train each model using a direct-block mean-squared-error objective, the same objective used in \cite{arndt2025synthetic}. Averaging over all \(F\) forecast states prevents the optimization from over-fixating on one position within the output window. It also provides a common objective that can be applied without modification to all recurrent, message-passing, attention, Transformer, and neural-operator architectures in the benchmark.

The direct-block loss is
\begin{equation}
\mathcal{L}_{\mathrm{direct}}
=
\frac{1}{F}
\sum_{\ell=1}^{F}
\left[
    \frac{1}{NC}
    \left\|
        \widehat{\widetilde{U}}^{(t+\ell)}
        -
        \widetilde{U}^{(t+\ell)}
    \right\|_{F}^{2}
\right].
\label{eq:direct-block-loss}
\end{equation}


\subsection{Benchmark datasets and evaluation protocol}
\label{sec:benchmark-datasets}

Models are optimized with AdamW \cite{loshchilov2017fixing} using the direct-block objective in Eq.~\eqref{eq:direct-block-loss}. Validation direct-block RMSE is monitored
during training, and the checkpoint with the best validation value is used for evaluation. All three benchmarks use scenario-level train, validation,
and test splits, with normalization statistics computed from the training partition only. When applicable across all three experiments, cosine
learning-rate decay is used; dataset-specific training budgets and learning rates are listed in Table~\ref{tab:benchmark-summary}.

Final performance is measured with the block-autoregressive procedure from
Section~\ref{sec:block-forecasting}, so model predictions are recursively
returned as future inputs after the initial observed history. We report mean
rollout RMSE over the full forecast horizon together with RMSE at the final
forecast state. Results are averaged over three independently trained runs.
Table~\ref{tab:benchmark-summary} displays the main dataset, forecasting,
model-capacity, and optimization settings for all three benchmarks.

\paragraph*{PFLOTRAN hydrologic surrogate}

The PFLOTRAN benchmark consists of three-dimensional coastal-groundwater simulations for a domain in Norfolk, Virginia \cite{hammond2014pflotran,liu2026hydrogeologic}. Each simulation evolves liquid pressure and salinity on a common unstructured subsurface graph.

\paragraph*{SI diffusion on a regional graph}

The SI-diffusion benchmark is generated from a susceptible--infected diffusion process over 400 European regions at level 3 of the Nomenclature of Territorial Units for Statistics (NUTS-3) classification \cite{arndt2025synthetic}. We use the infected-state field as the forecasting target.

\paragraph*{Airfoil computational fluid dynamics (CFD) surrogate}

The Airfoil benchmark is derived from the two-dimensional compressible-flow dataset of Pfaff \textit{et al.}~\cite{pfaff2021learning}, in which trajectories evolve on a common triangular airfoil mesh. The original simulations contain 600 time steps; we retain the first 200 states of each trajectory and predict density, pressure, and the two Cartesian velocity components.

\begin{table*}[t]
\centering
\caption{
Dataset, forecasting, and training settings for the three benchmarks.
All datasets use fixed graph topology within each benchmark, scenario-level
data partitions, and training-set normalization.
}
\label{tab:benchmark-summary}

\footnotesize
\setlength{\tabcolsep}{4pt}
\renewcommand{\arraystretch}{1.10}

\begin{tabularx}{0.96\textwidth}{
    @{}
    >{\raggedright\arraybackslash}p{0.22\textwidth}
    >{\centering\arraybackslash}X
    >{\centering\arraybackslash}X
    >{\centering\arraybackslash}X
    @{}
}
\toprule
\textbf{Setting}
&
\textbf{PFLOTRAN}
&
\textbf{SI diffusion}
&
\textbf{Airfoil CFD}
\\
\midrule

\multicolumn{4}{@{}l}{\textit{Dataset}} \\

Graph nodes
&
\(2{,}500\)
&
400
&
\(5{,}233\)
\\


Scenarios
&
100
&
25
&
\(1{,}200\)
\\

States per scenario
&
70
&
100
&
200
\\

Train / validation / test
&
80 / 10 / 10
&
18 / 3 / 4
&
\(1{,}000\) / 100 / 100
\\

\addlinespace[0.25em]
\midrule
\multicolumn{4}{@{}l}{\textit{Forecasting and model scale}} \\

History / forecast block
&
\(10 \rightarrow 10\)
&
\(14 \rightarrow 14\)
&
\(20 \rightarrow 20\)
\\

Rollout horizon
&
60
&
100
&
180
\\

Approximate capacity
&
\(1\)M
&
\(70\)k
&
\(10\)M
\\

\addlinespace[0.25em]
\midrule
\multicolumn{4}{@{}l}{\textit{Optimization}} \\

Maximum epochs
&
100
&
60
&
100
\\

Learning-rate range
&
\(5\times10^{-4}\) to \(10^{-6}\)
&
\(10^{-3}\) to \(5\times10^{-6}\)
&
\(10^{-4}\) to \( 2\times10^{-5}\)
\\

Early-stopping patience
&
10
&
15
&
None
\\

\bottomrule
\end{tabularx}
\end{table*}

\section*{Data availability}

The SI-diffusion dataset analyzed in this study is publicly available from the source dataset repository associated with Ref.~\cite{arndt2025synthetic}.
The processed benchmark data, scenario-level splits, and normalization information used for the experiments in this study are publicly available on
Hugging Face at
\url{huggingface.co/datasets/subaven/twig-benchmark-data}.

\section*{Code availability}

The source code for TWIG, capacity-matched baseline implementations,
training and evaluation configurations, and scripts used to reproduce the
experiments reported in this study will be made public upon publication at
\url{github.com/suba-ven/TWIG}.

\section*{Acknowledgements}

This work was prepared in partial fulfillment of the requirements of the Berkeley Lab Undergraduate Research (BLUR) Program, managed by the Academic Learning, Internships, and Faculty Training (A-LIFT) Office at Lawrence Berkeley National Laboratory. The research was supported by the Automated Scenario Assessment of Groundwater Table and Salinity Response to Sea-Level Rise project at Lawrence Berkeley National Laboratory, funded by the U.S. Department of Defense's Strategic Environmental Research and Development Program (SERDP). We also acknowledge Pacific Northwest National Laboratory and Stanford University for collaboration and support, funded in part by the Office of Advanced Scientific Computing Research (ASCR) within the Department of Energy Office of Science under award number DE-SC0023163.

\section*{Competing interests}
The authors declare no competing interests.

\bibliography{references}

\section{Appendix}

\subsection{Complete benchmark results}
\label{app:complete-results}

The main text emphasizes aggregate forecasting behavior and spatial error
structure. Tables~\ref{tab:pflotran-rollout},
\ref{tab:si-comparison}, and \ref{tab:Airfoil-comparison} report the
complete numerical comparison across all benchmark models. Within each
dataset, architectures use the same scenario-level splits, training
configuration, and block-autoregressive evaluation protocol. Results are
reported as mean \(\pm\) standard deviation across three independently
trained runs.

\begin{table*}[!htbp]
    \centering
    \caption{
        \textbf{Complete PFLOTRAN autoregressive forecasting results.}
        Mean rollout RMSE is averaged over forecast steps 1--60.
        Parameter counts are reported in millions.
    }
    \label{tab:pflotran-rollout}

    \small
    \setlength{\tabcolsep}{7pt}
    \renewcommand{\arraystretch}{1.10}

    \begin{tabular}{@{}lccc@{}}
        \toprule
        Method
        & Parameters (M)
        & Mean rollout RMSE
        & Final-timestep RMSE \\
        \midrule

        \textbf{TWIG (\(K=5,S=4\))}
        & 1.001
        & \(\mathbf{0.0136 \pm 0.0001}\)
        & \(\mathbf{0.0173 \pm 0.0004}\) \\

        GPS Transformer
        & 1.015
        & \(0.0140 \pm 0.0006\)
        & \(0.0177 \pm 0.0014\) \\

        GAT
        & 1.004
        & \(0.0143 \pm 0.0002\)
        & \(0.0182 \pm 0.0006\) \\

        MeshGraphNet
        & 0.976
        & \(0.0145 \pm 0.0001\)
        & \(0.0177 \pm 0.0005\) \\

        Graph WNO (\(S=4\))
        & 1.024
        & \(0.0145 \pm 0.0005\)
        & \(0.0185 \pm 0.0013\) \\

        GATv2
        & 0.995
        & \(0.0146 \pm 0.0001\)
        & \(0.0185 \pm 0.0007\) \\

        RNN--GNN Fusion
        & 0.974
        & \(0.0205 \pm 0.0003\)
        & \(0.0283 \pm 0.0017\) \\

        RNN
        & 0.995
        & \(0.0233 \pm 0.0009\)
        & \(0.0305 \pm 0.0021\) \\

        Graph FNO
        & 1.048
        & \(0.0545 \pm 0.0039\)
        & \(0.0833 \pm 0.0144\) \\

        \bottomrule
    \end{tabular}
\end{table*}

\begin{table*}[!htbp]
\centering
\caption{
    \textbf{Complete SI-diffusion autoregressive forecasting results.}
    Mean rollout RMSE is averaged over forecast steps 1--100.
    Parameter counts are reported in millions.
}
\label{tab:si-comparison}

\small
\setlength{\tabcolsep}{7pt}
\renewcommand{\arraystretch}{1.10}

\begin{tabular}{@{}lccc@{}}
\toprule
Method &
Parameters (M) &
Mean rollout RMSE &
Final-timestep RMSE \\
\midrule

\textbf{TWIG (\(K=5,S=4\))}
& 0.07023
& \(\mathbf{0.0576 \pm 0.0049}\)
& \(\mathbf{0.0470 \pm 0.0143}\) \\

Graph FNO
& 0.07086
& \(0.0602 \pm 0.0123\)
& \(0.0541 \pm 0.0050\) \\

GPS Transformer
& 0.06791
& \(0.0606 \pm 0.0111\)
& \(0.0701 \pm 0.0111\) \\

Graph WNO (\(S=4\))
& 0.06420
& \(0.0617 \pm 0.0089\)
& \(0.0612 \pm 0.0163\) \\

GATv2
& 0.06945
& \(0.0664 \pm 0.0057\)
& \(0.0574 \pm 0.0058\) \\

GAT
& 0.06751
& \(0.0673 \pm 0.0027\)
& \(0.0581 \pm 0.0076\) \\

MeshGraphNet
& 0.06901
& \(0.0751 \pm 0.0080\)
& \(0.0744 \pm 0.0144\) \\

RNN--GNN Fusion
& 0.07225
& \(0.1002 \pm 0.0018\)
& \(0.0802 \pm 0.0002\) \\

RNN
& 0.06785
& \(0.1250 \pm 0.0200\)
& \(0.0994 \pm 0.0226\) \\

\bottomrule
\end{tabular}
\end{table*}

\begin{table*}[!htbp]
\centering
\caption{
    \textbf{Complete Airfoil autoregressive forecasting results.}
    Mean rollout RMSE is averaged over forecast steps 1--180.
    The main table reports the best tested graph-wavelet spatial-scale
    configuration for each graph-wavelet architecture:
    \(S=12\) for TWIG and \(S=4\) for Graph WNO.
}
\label{tab:Airfoil-comparison}

\small
\setlength{\tabcolsep}{6pt}
\renewcommand{\arraystretch}{1.10}

\begin{tabular}{@{}lccc@{}}
\toprule
Method &
Parameters (M) &
Mean rollout RMSE &
Final-timestep RMSE \\
\midrule

GPS Transformer
& 9.85
& \(\mathbf{0.1038 \pm 0.0080}\)
& \(\mathbf{0.1856 \pm 0.0299}\) \\

TWIG (\(K=5,S=12\))
& 9.88
& \(0.1563 \pm 0.0035\)
& \(0.2980 \pm 0.0152\) \\

MeshGraphNet
& 10.04
& \(0.1909 \pm 0.0104\)
& \(0.3303 \pm 0.0107\) \\

Graph WNO (\(S=4\))
& 10.03
& \(0.2011 \pm 0.0042\)
& \(0.3276 \pm 0.0054\) \\

Graph FNO
& 10.00
& \(0.2026 \pm 0.0066\)
& \(0.3363 \pm 0.0185\) \\

GAT
& 9.99
& \(0.2755 \pm 0.0150\)
& \(0.5438 \pm 0.0566\) \\

GATv2
& 9.97
& \(0.2762 \pm 0.0120\)
& \(0.4994 \pm 0.0129\) \\

RNN--GNN Fusion
& 9.97
& \(0.7130 \pm 0.1198\)
& \(1.5502 \pm 0.3634\) \\

RNN
& 9.92
& \(0.8922 \pm 0.0854\)
& \(1.3925 \pm 0.1375\) \\

\bottomrule
\end{tabular}
\end{table*}

\FloatBarrier

\subsection{Benchmark architecture details}
\label{app:baseline-details}

All three benchmarks use controlled, capacity-matched adaptations of the same
architecture families. Target model capacities are approximately \(70\)k
parameters for SI diffusion, \(1\)M for PFLOTRAN, and \(10\)M for Airfoil.
Table~\ref{tab:model-configurations} reports the parameter counts and
multiscale settings of the configurations used in the main benchmark
comparisons. Additional temporal- and spatial-scale variants are evaluated
separately in the ablation experiments.

\begin{table*}[!htbp]
\centering
\caption{
\textbf{Capacity-matched model configurations across the three benchmarks.}
Parameter counts are reported in millions. Parenthetical values denote the
number of time-causal temporal bands \(K\) and graph-wavelet spatial scales
\(S\), where applicable.
}
\label{tab:model-configurations}

\small
\setlength{\tabcolsep}{5pt}
\renewcommand{\arraystretch}{1.15}

\begin{tabular}{@{}lccc@{}}
\toprule
\textbf{Model} &
\textbf{SI diffusion} &
\textbf{PFLOTRAN} &
\textbf{Airfoil} \\
\midrule

TWIG
& \(0.07023\;(K=5,S=4)\)
& \(1.001\;(K=5,S=4)\)
& \(9.88\;(K=5,S=12)\) \\

Graph WNO
& \(0.06420\;(S=4)\)
& \(1.024\;(S=4)\)
& \(10.03\;(S=4)\) \\

Graph FNO
& \(0.07086\)
& \(1.048\)
& \(10.00\) \\

MeshGraphNet
& \(0.06901\)
& \(0.976\)
& \(10.04\) \\

GAT
& \(0.06751\)
& \(1.004\)
& \(9.99\) \\

GATv2
& \(0.06945\)
& \(0.995\)
& \(9.97\) \\

GPS Transformer
& \(0.06791\)
& \(1.015\)
& \(9.85\) \\

RNN
& \(0.06785\)
& \(0.995\)
& \(9.92\) \\

RNN--GNN Fusion
& \(0.07225\)
& \(0.974\)
& \(9.97\) \\

\bottomrule
\end{tabular}
\end{table*}

\subsection{Ablation and robustness configurations}
\label{app:ablation-configurations}

The main ablation experiments vary one component of the multiscale
representation while retaining the remaining benchmark configuration.
Table~\ref{tab:ablation-configurations} summarizes the tested settings.

\begin{table}[!htbp]
\centering
\caption{
\textbf{Configurations used for the multiscale and robustness experiments.}
}
\label{tab:ablation-configurations}

\small
\renewcommand{\arraystretch}{1.12}
\setlength{\tabcolsep}{6pt}

\begin{tabular}{@{}lll@{}}
\toprule
Experiment &
Dataset &
Tested settings \\
\midrule

Temporal resolution
& PFLOTRAN
& \(K\in\{3,5,7\}\) \\

Spatial resolution
& Airfoil
& \(S\in\{4,8,12\}\) \\

Initial-history noise
& PFLOTRAN
& \(0,1,2.5,5,10\%\) \\

\bottomrule
\end{tabular}
\end{table}

For the temporal-resolution experiment, model widths are adjusted to maintain
approximately \(1\)M trainable parameters as \(K\) changes. For the
spatial-resolution experiment, TWIG and Graph WNO are compared at the same
tested values of \(S\), with model capacity maintained near \(10\)M
parameters. In the perturbation experiment, noise is applied only to the
observed states used to initialize the rollout and is scaled independently by
each channel's training-set standard deviation. Models are trained only on
clean data.

\subsection{Relationship to source architectures}
\label{app:source-architecture-relationship}

To match parameter capacity within each benchmark, the baselines are
controlled adaptations of established architecture families rather than exact
reproductions of their original training pipelines. All models use the same
forecasting targets, scenario-level data splits, normalization, direct-block
training objective, and autoregressive evaluation procedure within each
benchmark.

\paragraph*{TWIG and Graph WNO.}

The graph-wavelet models adapt wavelet neural operators from regular domains
to irregular graphs~\cite{tripura2023wavelet}. Instead of applying a
regular-grid wavelet transform, the spatial operator defines multiscale
spectral filters in a truncated graph-Laplacian basis. Plain Graph WNO
receives the raw state history directly as input channels and does not use
the time-causal multiscale temporal encoder. TWIG first transforms the
history into causal temporal residual and memory features before
graph-wavelet propagation.

\paragraph*{Graph FNO.}

Graph FNO is the corresponding graph-spectral analogue of the Fourier neural
operator~\cite{li2021fourier}. It replaces fast Fourier transform modes with
graph-Laplacian eigenmodes and performs learned spectral mixing in the
truncated graph basis. It uses the raw state history and does not include the
time-causal temporal encoder.

\paragraph*{MeshGraphNet.}

MeshGraphNet retains the encode--process--decode message-passing structure of
Pfaff \textit{et al.}~\cite{pfaff2021learning}. Learned edge updates are
aggregated into node updates through a sequence of residual processor blocks,
followed by a decoder that predicts the complete future-state block. Our
implementation is capacity matched within each benchmark and evaluated under
the common block-autoregressive forecasting protocol.

\paragraph*{GAT and GATv2.}

GAT and GATv2 preserve the neighborhood-attention mechanisms of their source
architectures~\cite{velickovic2018graph,brody2022how}. The attention layers
are incorporated into residual graph-field forecasting networks with
pointwise feed-forward channel mixing and a block-output decoder. Model widths
and attention dimensions are adjusted within each benchmark to approximately
match the target parameter capacity.

\paragraph*{GPS Transformer.}

The GPS Transformer follows the GraphGPS design principle of combining local
message passing with global multi-head self-attention and pointwise
feed-forward updates~\cite{rampasek2022recipe}. Our implementation uses the
same graph-valued history and static structural information supplied to the
other spatial baselines and jointly predicts the future-state block.

\paragraph*{RNN.}

The recurrent baseline uses gated recurrent unit (GRU) recurrence
\cite{cho2014learning} independently at each graph node. It therefore models
temporal evolution without explicit graph message passing and serves as a
temporal-only control.

\paragraph*{RNN--GNN Fusion.}

The RNN--GNN Fusion baseline combines a GRU-based temporal branch with a graph
message-passing branch through a learned fusion mechanism before block
decoding. It follows the recurrent--graph model family used in the
SI-diffusion benchmark~\cite{arndt2025synthetic}, while being capacity
matched and evaluated under the common forecasting protocol used here.
\end{document}